\documentclass[runningheads]{llncs}
\usepackage{graphicx}
\usepackage{graphics}
\usepackage{booktabs}
\usepackage{graphicx,epstopdf,subfigure}
\usepackage{amsmath,amssymb,bm,bbm}
\usepackage{algorithmic}
\usepackage{tabularx,array,booktabs}
\usepackage{cite}
\usepackage{caption}
\usepackage{comment}
\usepackage{multirow}
\usepackage{url}
\usepackage{color}
\usepackage{subfigure}
\usepackage{amssymb}
\usepackage{caption}
\makeatletter
\newcommand*{\rom}[1]{\expandafter\@slowromancap\romannumeral #1@}
\makeatother

\begin{document}
\title{Unifying Structure Analysis and Surrogate-driven Function Regression for Glaucoma OCT Image Screening}
\titlerunning{Structure Analysis and Function Regression for Glaucoma OCT Screening}
\author{Xi Wang\inst{1}\and
	Hao Chen\inst{2}\thanks{Corresponding author} \and Luyang Luo\inst{1} \and 
	An-ran Ran\inst{3} \and Poemen P. Chan\inst{3}, \\ Clement C. Tham\inst{3} \and Carol Y. Cheung\inst{3} \and Pheng-Ann Heng\inst{1,4} \\
	\email{\{xiwang,hchen\}@cse.cuhk.edu.hk}
	}
\authorrunning{X. Wang et al.}
\institute{Department of Computer Science and Engineering,\\ The Chinese University of Hong Kong, Hong Kong,
	China \and
Imsight Medical Technology, Co., Ltd., China \and
Department of Ophthalmology and Visual Sciences,\\ The Chinese University of Hong Kong, Hong Kong,
China \and
Guangdong Provincial Key Laboratory of Computer Vision and Virtual Reality Technology, Shenzhen Institutes of Advanced Technology, Chinese Academy of Sciences, Shenzhen, China \\
}

\maketitle             

\begin{abstract}
Optical Coherence Tomography (OCT) imaging plays an important role in glaucoma diagnosis in clinical practice. Early detection and timely treatment can prevent glaucoma patients from permanent vision loss. However, only a dearth of automated methods has been developed based on OCT images for glaucoma study. In this paper, we present a novel framework to effectively classify glaucoma OCT images from normal ones. A semi-supervised learning strategy with \textit{smoothness assumption} is applied for surrogate assignment of missing function regression labels. Besides, the proposed multi-task learning network is capable of exploring the structure and function relationship from the OCT image and visual field measurement simultaneously, which contributes to classification performance boosting. Essentially, we are the first to unify the structure analysis and function regression for glaucoma screening. It is also worth noting that we build the largest glaucoma OCT image dataset involving 4877 volumes to develop and evaluate the proposed method. Extensive experiments demonstrate that our framework outperforms the baseline methods and two glaucoma experts by a large margin, achieving 93.2\%, 93.2\% and 97.8\% on accuracy, F1 score and AUC, respectively.

\end{abstract}

\section{Introduction}

Glaucoma is the most frequent cause of irreversible blindness worldwide, which is a heterogeneous group of disease that damages the optic nerve and leads to vision loss~\cite{JONAS20172183}. 
It is featured by loss of retinal ganglion cells, thinning of retinal nerve fibre layer (RNFL), and cupping of the optic disc. 
Early detection and diagnosis are essential as they can facilitate immediate treatment and prevent the progression of the disease, especially for early-stage glaucoma patients. On clinical grounds, diagnosis of glaucoma usually requires a variety of tests, including functional measurement (e.g., visual field test)
 and structure assessment (e.g., optical coherence tomography). 
In general, visual field (VF) test provides three important global indices: visual field index (VFI), mean deviation (MD) and pattern standard deviation (PSD) to depict the functional changes.
Optical coherence tomography (OCT) is a non-contact and non-invasive imaging modality that generates high-resolution, cross-sectional images (B-scans) of the retina. It can provide objective and quantitative assessment of various retinal structures. 
However, the examination of OCT imaging requires highly trained ophthalmologist, which is always partly subjective and time-consuming. Thus, automated glaucoma OCT image screening tool is of great needs in clinical practice. Moreover, evaluation of the relationship between structural and functional damage can provide valuable insight into how visual function works according to the degree of structural damage, which can help our understanding of glaucoma.
This is quite significant in diagnosing, staging and monitoring glaucomatous patients.

Many researchers have devoted their efforts to solving the glaucoma OCT image screening problem and have achieved significant progress. Nevertheless, most of the previous works were based on machine learning methods and heavily relied on established features, such as the measurements on retinal nerve fiber layer thickness and ganglion cell layer thickness~\cite{huang2005development,kim2016glaucoma,christopher2018retinal}.
Differently, Ramzan et al.~\cite{ramzan2018automated} proposed an approach that first segmented inner limiting membrane and retinal pigment epithelium layers in OCT images for cup-to-disc ratio calculation, and then differentiated glaucoma based on the calculated ratio. 
A pioneering work~\cite{maetschke2018feature} recently proposed a 3D convolutional neural network (CNN) to directly classify the downsampled OCT volumes into glaucoma or normal, which considerably outperformed various feature-based machine learning algorithms. 
With respect to modelling the structure-function relationship\cite{el2003retinal,leite2012structure}, Leite et al. utilized locally weighted scatterplot smoothing and regression analysis to evaluate parapapillary RNFL thickness sectors and corresponding topographic standard automated perimetry locations~\cite{leite2012structure}.
However, there are still three main challenges that have not been fully investigated yet. First, heretofore there is a dearth of studies that explores the structure and function relationship based on the raw OCT image and visual field measurement for glaucoma screening. Second, owing to very limited images in current datasets, validation experiments of previous methods were not comprehensive, which indeed constrained the development of robust and reliable approaches. Third, building a large medical dataset is always confronted with many difficulties, e.g., it is inevitable to acquire incomplete clinical records due to some unexpected reasons, which is a common phenomenon in retrospective studies.

In this paper, we carried thorough investigation on all of the aforementioned challenges. To the best of our knowledge, we are the first to unify structure analysis and function regression for glaucoma screening based on OCT images. We develop a novel framework that explores the structure and function relationship between OCT image and visual field measurement via a semi-supervised multi-task learning network. Specifically, a semi-supervised learning method is first introduced to find the surrogates to fill the vacancy of missing visual field measurements. Afterwards, both class labels (glaucoma/normal) and visual field measurements (ground truth and pseudo labels) are utilized to jointly train a multi-task learning network. In particular, we concatenate the classification features with those learned from the regression module to identify glaucoma. It indicates that the structure and function relationship learned by our network indeed contributes to classification performance improvement.
It is worthwhile to emphasize that to our best knowledge, the largest glaucoma OCT　image dataset composed of $ 4877 $ volumes of the optic disc is constructed in this study for algorithm development and evaluation.

\section{Method}
\begin{figure}
	\center
	\includegraphics[width=0.95\linewidth]{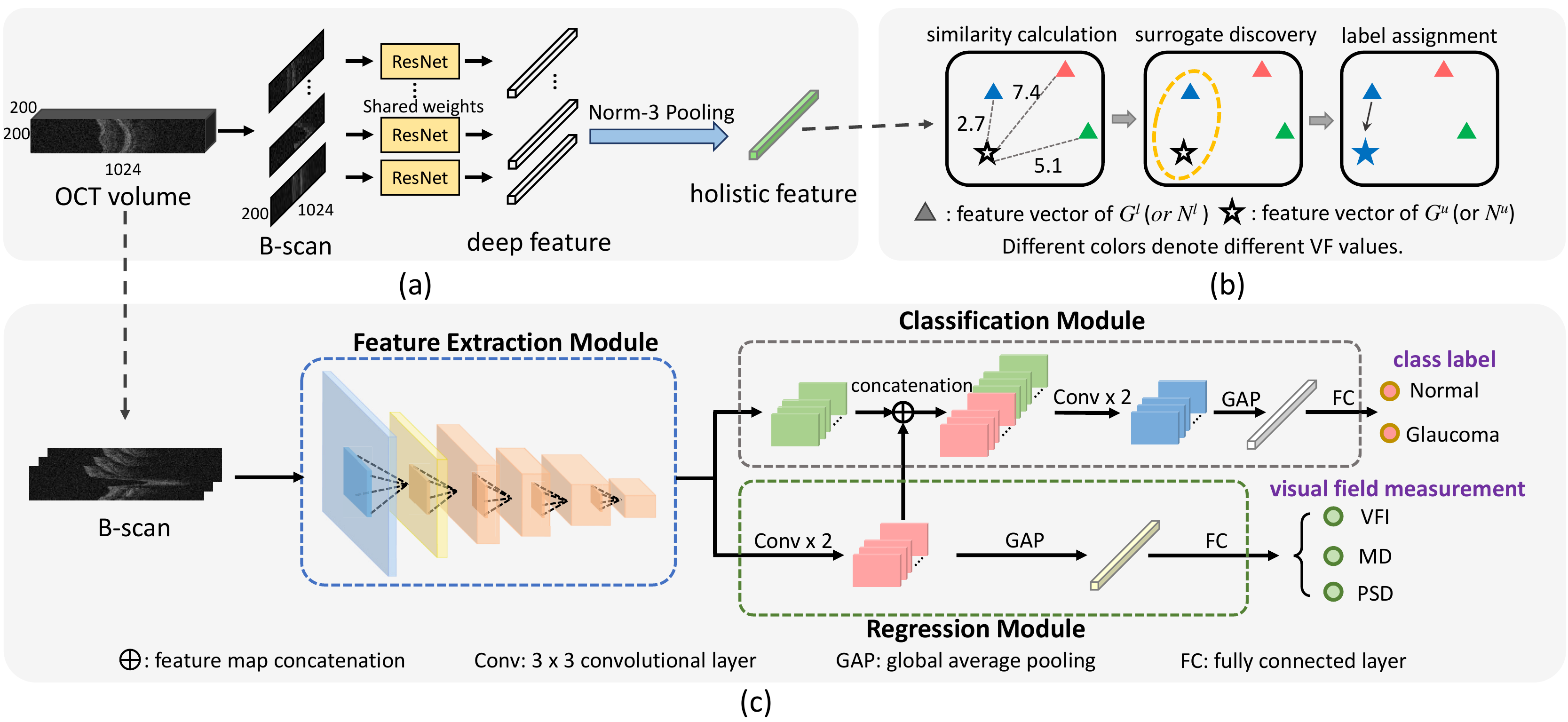}
	\caption{An overview of the proposed method. (a) Initially, we train a CNN model with class labels (glaucoma/normal) to extract features of B-scans. Then the extracted features are aggregated to form a global representation of an OCT volume. (b) Next, we compute similarities between homogeneous groups to find the nearest neighbour of OCT images without VF measurement and enable its VF values for surrogate assignment. Here, $G^{l}$ and $G^{u}$ indicate glaucoma images with and without VF labels while $N^{l}$ and $N^{u}$ denote normal images with and without VF labels. (c) Lastly, class labels and VF labels (ground truth and pseudo labels) are unified to train a multi-task learning network.
	}
	\label{fig:framework}
\end{figure}

Our main goal is to effectively classify glaucoma OCT images assisted by exploring the structure and function relationship of glaucoma.
Fig.\ref{fig:framework} illustrates the overview of the proposed method, which consists of two parts. The first part uses a semi-supervised learning technique to solve the missing label problem for function regression. In particular, a B-scan-based CNN is trained under full supervision of class labels, aiming to extract the holistic representation for each OCT volume. Afterwards, pseudo labels are automatically generated by probing the nearest neighbour of OCT images without VF measurement among homogeneous groups with CNN-encoded features. In the second part, a multi-task learning network is trained to unify structure analysis and surrogate-driven function regression for more accurate glaucoma screening.

\subsection{Surrogate-driven Labelling with Semi-supervised Learning}
In order to solve the problem of missing VF measurement labels, we borrow the spirit from semi-supervised learning and come up with an appropriate solution.
In semi-supervised learning domain, the \textit{smoothness assumption} points out that features close to each other are more likely to share the same label. This assumption is intimately linked to a definition of what it means for one feature to be near another feature, which can be embodied in a similarity function $S(\cdot,\cdot)$ on input space~\cite{chapelle2009semi}. 

To project OCT volumes into the input space for similarity calculation, we adopted ResNet18~\cite{he2016deep} as the feature representation model. Specifically, this CNN is first trained based on B-scans under full supervision of class labels. Then we apply the trained model to extract the embedded feature of each B-scan from the Global Average Pooling (GAP) layer. Finally following\cite{wang2018weakly}, the norm-3 pooling, $\mathcal{F}=(\sum_{i=1}^{n}f_{i}^{3})^{\frac{1}{3}}$, is applied to aggregate B-scan features into a holistic representation $ \mathcal{F} $ for each OCT volume. Here, $f$ is the feature representation of B-scan input, and $n$ is the number of B-scans from the same volume.

Based on the availability of class labels and the VF measurements, we group the training set into four clusters: (1) $G^{l}$:  glaucoma images with VF measurement; (2) $G^{u}$: glaucoma images without VF measurement; (3) $N^{l}$: normal images with VF measurement and (4) $N^{u}$: normal images without VF measurement. 
Through the feature representation model, each OCT volume is represented by a feature vector, $\mathcal{F} \in\mathbb{R}^{512}$. 
The similarity between any pair of homogeneous OCT images, e.g., $\mathcal{F}^{l}\in G^{l}$ and $\mathcal{F}^{u}\in G^{u}$, is measured by \textit{Euclidean distance}:

\begin{equation}\label{Eq1}
S(\mathcal{F}^{l},{\mathcal{F}^{u}})=[{\sum_{i=1}^{m}(\mathcal{F}_{i}^{l}-\mathcal{F}_{i}^{u})^{2}}]^{\frac{1}{2}}
\end{equation}

\indent
Inspired by the semi-supervised \textit{smoothness assumption}, for each OCT image without VF measurement $\mathcal{F}_{j}^{u}\in G^{u}$ (or $\mathcal{F}_{j}^{u}\in N^{u}$), we find its nearest neighbour in the group of the same class that has VF measurement $\mathcal{F}_{i^{*}}^{l}\in G^{l}$ (or $\mathcal{F}_{i^{*}}^{l}\in N^{l}$), where $i^{*}=\mathop{\arg\min}_{i}{S(\mathcal{F}_{i}^{l},\mathcal{F}_{j}^{u})}$, and appoint the VF measurement of $\mathcal{F}_{i^{*}}^{l}$ to $\mathcal{F}_{j}^{u}$. Consequently, we successfully find suitable surrogates for all missing VF measurements in this semi-supervised learning fashion.

\subsection{Multi-task Learning for Structure and Function Analysis}
We proposed an end-to-end multi-task learning CNN with a primary task to classify B-scan images into glaucomatous and normal, and an auxiliary task to investigate the relationship between structural and functional changes of glaucoma eyes. 
Particularly, this network is composed of three components, the shared feature extraction module, the classification module and the regression module. As illustrated in Fig.\ref{fig:framework}(c), we employ ResNet18, initialized with ImageNet pretrained weights, as the backbone in the feature extraction module whose weights are shared by both classification and regression tasks. The rest of the network is composed of two branches, one for glaucoma discrimination and the other for visual field measurement regression. These tasks are trained jointly.

\subsubsection{Visual Field Measurement Regression.}
For each attribute of visual field measurement, i.e., VFI, MD and PSD, we formulate it as an individual regression task. Within the regression module, two convolutional layers with ReLU activation and batch-normalization are inserted before GAP. Fully connected layers with sigmoid activation are then used to regress these attributes. The regression tasks are driven by minimizing the mean square error:

\begin{equation}\label{Eq5}
\mathcal{L}_{reg}=\dfrac{1}{N}\sum_{i=1}^{N}\lVert{y_{i}^{r}}-\hat{y}_{i}^{r}(x_{i};\theta_{s},\theta_{reg})\rVert_{2}^{2}
\end{equation}
\noindent
where $ N $ denotes the number of training samples, $ x_{i} $ is the input of three adjacent B-scan images, $ y_{i}^{r} $ is the clinical measurement of visual field attribute and $\hat{y}_{i}^{r} $ is the corresponding prediction of the network. $\theta_{s} $ denotes shared weights in the feature extraction module and $\theta_{reg}$ denotes features in the regression module, respectively. In this paper, we use superscripts $r$ and $c$ for discrimination between \textbf{r}egression and  \textbf{c}lassification tasks.

\subsubsection{Glaucoma Classification.}
The classification module performs the primary task for glaucoma screening. In routine clinical practice, visual field measurement is an important indicator of functional change for glaucoma diagnosis. Hence, it is reasonable to hypothesize that if the relationship between structure and function is appropriately discovered, the learned features in the regression module could exert a positive influence on the classification task.
Based on this assumption, we concatenate the attribute regression feature maps with those originated from the feature extraction module, supposing that the features learned in the regression module could provide the classifier with helpful guidance. Here, the typical binary cross-entropy loss is utilized to train this classifier:
\begin{equation}\label{Eq6}
\mathcal{L}_{cls}=-\dfrac{1}{N}\sum_{i=1}^{N}y_{i}^{c}\log \hat{y}_{i}^{c}(x_{i};\theta_{s},\theta_{cls})
\end{equation}
\noindent
where $ y_{i}^{c} $ is the ground truth label while $ \hat{y}_{i}^{c} $ is the likelihood predicted by the classifier.
Similarly, $\theta_{cls}$ stands for weights in the classification module.

\subsubsection{Joint Training of Multi-task Learning Network.}
Finally, the multi-task learning network is trained by minimizing the weighted combination of the mean square error losses and the binary cross-entropy loss.

\begin{equation}\label{Eq7}
\mathcal{L}_{total}=\mathcal{L}_{cls}+\sum_{j=1}^{3}\alpha_{j}\mathcal{L}_{reg}^{j}
\end{equation}
\noindent
where $\alpha_{j}$ is the hyper-parameter balancing $\mathcal{L}_{cls}$ and $\mathcal{L}_{reg}^{j}$ determined by cross validation.
At the inference stage, we take the average of multiple B-scan-level probabilities as a single volume-level prediction.

\section{Experiments and Results}
\textbf{Dataset:}
In this study, we constructed the largest scale cohort for evaluation. This dataset consists of $4877$ volumetric OCT images (glaucoma: 2926; normal: 1951) from $930$ subjects. It is worth noting that each investigated subject (one eye or both two eyes involved) might have several follow-ups, and also several OCT images may be taken during each follow-up, which eventually results in $ 3182 $ eye visits (glaucoma: 1901; normal: 1281) in total. Specially, we denote the eye-visit result as \textit{case-level} result.
A part of VF measurements for some follow-ups are unavailable in this study.
Two glaucoma specialists worked individually to label all the OCT images into glaucoma and normal, taking VF tests and other clinical records as reference. A senior glaucoma expert was consulted in case of disagreement. 
Subsets of $2895$, $1015$ and $967$ images are randomly selected for training, validation and testing, respectively. The random sampling is at patient level so as to prevent leakage and biased estimation of the testing performance. According to the accessibility of VF measurements in the training set, we re-configure the training set as follows:
\noindent
(\romannumeral1). \textit{Part}: $1979$ images whose VF measurements exist.
\noindent
(\romannumeral2). \textit{All}: all $2895$ images. 

\noindent
\textbf{Quantitative Evaluation and Comparison:} The performance is measured via three criteria: classification accuracy, F1 score and Area Under ROC Curve (AUC).
To obtain the \textit{case-level} prediction, averaging method is used to aggregate the results of several images during each eye visit to a single one. 

At present, there is only a dearth of studies on glaucoma OCT image classification. Hence, several baseline methods, including  an existing method as well as three variants of the proposed method, were implemented for comparison: 
(\romannumeral1). 3D-CNN: the implementation of the approach proposed in~\cite{maetschke2018feature} which is trained with downsampled 3D volumes.
(\romannumeral2). 3D-ResNet: A 3D implementation of ResNet18~\cite{he2016deep} that takes raw volumes as input.
(\romannumeral3). 2D-ResNet: ResNet18 trained with B-scan images from OCT volumes. 
(\romannumeral4). 2D-ResNet-MT: 2D-ResNet with Multi-Task learning network as shown in Fig.\ref{fig:framework}(c) without surrogate label assignment. Specifically, if the training sample is lack of visual field measurement, the regression loss is ignored.
(\romannumeral5). 2D-ResNet-SEMT: The proposed SEmi-supervised Multi-Task learning network with surrogate label assignment.

The experimental results are listed in Table~\ref{Tab1}.
At first, trained with the same set \textit{All}, 2D-ResNet is superior to 3D-ResNet. There are two possible reasons. One is that training a 3D network with such high-dimensional input is extremely difficult. In fact, validation loss oscillates wildly during training, which makes it hard
\begin{table}
	\centering
	\caption{Comparison with different methods and expert performance.}
	\label{Tab1}
	\resizebox{11cm}{!}{%
		\begin{tabular}{l|l|cccccc}
			\hline
			\multicolumn{1}{l|}{\multirow{2}{*}{\multirow{2}{*}{Data}}}                  &\multicolumn{1}{l|}{\multirow{2}{*}{\multirow{2}{*}{Methods}}}                  & \multicolumn{2}{c}{Accuracy} & \multicolumn{2}{c}{F1 Score} & \multicolumn{2}{c}{AUC}  \\
			\multicolumn{1}{l|}{}    &\multicolumn{1}{l|}{}  & image-level   & case-level   & image-level   & case-level   & image-level & case-level \\
			\hline
			\multirow{3}{*}{\textit{Part}} 
			& 2D-ResNet           & 0.823         & 0.829        & 0.835         & 0.830        & 0.960       & 0.955      \\
			& 2D-ResNet-MT        & 0.878         & 0.882        & 0.890         & 0.887        & 0.971       & 0.968      \\     
			\hline
			\multirow{4}{*}{\textit{All}}
			&3D-CNN~\cite{maetschke2018feature}& 0.884          & 0.889          & 0.881          & 0.890          & 0.962          & 0.959          \\
			&3D-ResNet           & 0.880          & 0.875          & 0.878          & 0.874          & 0.958          & 0.956          \\
			&2D-ResNet           & 0.908          & 0.911          & 0.904          & 0.909          & 0.968          & 0.964          \\
			&2D-ResNet-MT           & 0.915          & 0.912          & 0.923          & 0.917          & 0.975          & 0.971          \\
			&2D-ResNet-SEMT & \textbf{0.932} & \textbf{0.924} & \textbf{0.932} & \textbf{0.924} & \textbf{0.978} & \textbf{0.973} \\ 
			\hline
			
			& Expert 1 & 0.912         & 0.912        & 0.917         & 0.917        & 0.918       & 0.918      \\
			& Expert 2 & 0.905         & 0.905        & 0.913         & 0.913        & 0.914       & 0.914      \\
			\hline
		\end{tabular}%
	}
\end{table}
for selecting models. The other is that there is a deficiency of 3D pre-trained model available, thus training from scratch could readily result in local optima.
By exploring the structure and function relationship in 2D-ResNet-MT, the classification performance is improved, which verifies our hypothesis that the extra information from the regression module is helpful.  
Inspiringly, the proposed framework achieves the best results among the aforementioned methods on all metrics. With surrogate label assignment for VF measurement, the classification module can receive more reliable information from the regression branch.

When compared to the pioneering work 3D-CNN~\cite{maetschke2018feature}, our method outperforms it by a large margin, with $4.8\%$, $5.1\%$, and $1.6\%$ performance improvement on accuracy, F1 score and AUC at image level. By and large, 3D-CNN has two drawbacks. First, the input volumes are compressed intensively, which may lead to discriminative information loss. Second, its capability is quite limited due to the shallow network structure.
To further explore the efficacy of our method, we invited another two glaucoma experts to identify glaucoma based on the printout of the OCT images in the format that ophthalmologists usually read in clinic. They reviewed the printouts individually masked from any other clinical notes to make the decision, either glaucoma or normal, for each testing image. Apparently, the proposed method exceeds expert performance significantly, particularly on AUC.

\begin{figure}[t]
	
	\center
	\includegraphics[width=1\linewidth]{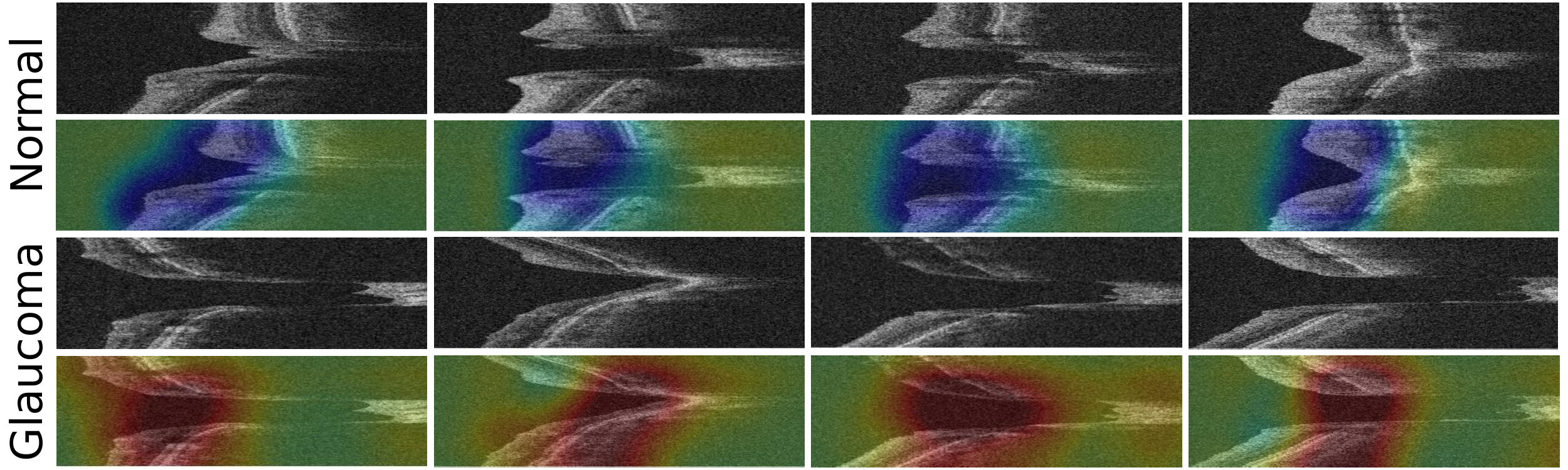}
	\caption{Visualization of discriminative regions. The first two rows show the pairs of normal B-scan and corresponding CAM. The last two rows show the pairs of glaucomatous ones. (Best viewed in color) }
	\label{fig:heatmap}

\end{figure}
\noindent
\textbf{Qualitative Evaluation} Class Activation Maps (CAMs)~\cite{zhou2016learning} were computed to visualize the discriminative regions that played a vital role in class prediction. For each pair in Fig.\ref{fig:heatmap}, the upper one is the input, and the lower one is the corresponding CAM overlapped with the input. 
Clearly, the discriminative regions found by CNN for glaucoma and normal images are quite different. In normal B-scan images, there is barely any response within retinal areas while such regions contrarily have high responses in glaucoma images. It is corresponding with the clinical diagnosis of glaucoma.
\section{Conclusion}
In this study, we present a deep learning framework to screen glaucoma based on volumetric OCT images of the optic disc. We first use a semi-supervised learning method to address the problem of incomplete visual field measurement labels of training images. Then a multi-task learning network is built to explore the relationship between functional and structural changes of glaucoma, which is verified beneficial to classification improvement. Extensive experiments on our large-scale dataset manifested the effectiveness of the proposed approach, which outperforms the baseline methods by a large margin. In addition, the comparison with glaucoma specialists provides strong evidence that our proposed framework has promising potential for automated glaucoma screening in the near future.

\subsubsection{Acknowledgements}
This project is supported in part by the National Basic Program of China 973 Program under Grant 2015CB351706, grants from the National Natural Science Foundation of China with Project No. U1613219, Research Grants Council - General Research Fund, Hong Kong (Ref: 14102418) and Shenzhen Science and Technology Program (No. JCYJ20180507182410327).

%
\bibliographystyle{IEEEtran}
\bibliography{IEEEfull}

\begin{thebibliography}{10}
\providecommand{\url}[1]{#1}
\csname url@samestyle\endcsname
\providecommand{\newblock}{\relax}
\providecommand{\bibinfo}[2]{#2}
\providecommand{\BIBentrySTDinterwordspacing}{\spaceskip=0pt\relax}
\providecommand{\BIBentryALTinterwordstretchfactor}{4}
\providecommand{\BIBentryALTinterwordspacing}{\spaceskip=\fontdimen2\font plus
\BIBentryALTinterwordstretchfactor\fontdimen3\font minus
  \fontdimen4\font\relax}
\providecommand{\BIBforeignlanguage}[2]{{%
\expandafter\ifx\csname l@#1\endcsname\relax
\typeout{** WARNING: IEEEtran.bst: No hyphenation pattern has been}%
\typeout{** loaded for the language `#1'. Using the pattern for}%
\typeout{** the default language instead.}%
\else
\language=\csname l@#1\endcsname
\fi
#2}}
\providecommand{\BIBdecl}{\relax}
\BIBdecl

\bibitem{JONAS20172183}
J.~B. Jonas, T.~Aung, R.~R. Bourne, A.~M. Bron, R.~Ritch, and S.~Panda-Jonas,
  ``Glaucoma,'' \emph{The Lancet}, vol. 390, pp. 2183--2193, 2017.

\bibitem{huang2005development}
M.-L. Huang and H.-Y. Chen, ``Development and comparison of automated
  classifiers for glaucoma diagnosis using stratus optical coherence
  tomography,'' \emph{Investigative ophthalmology \& visual science}, vol.~46,
  no.~11, pp. 4121--4129, 2005.

\bibitem{kim2016glaucoma}
H.~J. Kim, S.-Y. Lee, K.~H. Park, D.~M. Kim, and J.~W. Jeoung, ``Glaucoma
  diagnostic ability of layer-by-layer segmented ganglion cell complex by
  spectral-domain optical coherence tomography,'' \emph{Investigative
  ophthalmology \& visual science}, vol.~57, no.~11, pp. 4799--4805, 2016.

\bibitem{christopher2018retinal}
M.~Christopher, A.~Belghith, R.~N. Weinreb, C.~Bowd, M.~H. Goldbaum, L.~J.
  Saunders, F.~A. Medeiros, and L.~M. Zangwill, ``Retinal nerve fiber layer
  features identified by unsupervised machine learning on optical coherence
  tomography scans predict glaucoma progression,'' \emph{Investigative
  ophthalmology \& visual science}, vol.~59, no.~7, pp. 2748--2756, 2018.

\bibitem{ramzan2018automated}
A.~Ramzan, M.~U. Akram, A.~Shaukat, S.~G. Khawaja, U.~U. Yasin, and W.~H. Butt,
  ``Automated glaucoma detection using retinal layers segmentation and optic
  cup-to-disc ratio in optical coherence tomography images,'' \emph{IET Image
  Processing}, 2018.

\bibitem{maetschke2018feature}
S.~Maetschke, B.~Antony, H.~Ishikawa, and R.~Garvani, ``A feature agnostic
  approach for glaucoma detection in oct volumes,'' \emph{arXiv preprint
  arXiv:1807.04855}, 2018.

\bibitem{el2003retinal}
T.~A. El~Beltagi, C.~Bowd, C.~Boden, P.~Amini, P.~A. Sample, L.~M. Zangwill,
  and R.~N. Weinreb, ``Retinal nerve fiber layer thickness measured with
  optical coherence tomography is related to visual function in glaucomatous
  eyes,'' \emph{Ophthalmology}, vol. 110, no.~11, pp. 2185--2191, 2003.

\bibitem{leite2012structure}
M.~T. Leite, L.~M. Zangwill, R.~N. Weinreb, H.~L. Rao, L.~M. Alencar, and F.~A.
  Medeiros, ``Structure-function relationships using the cirrus spectral domain
  optical coherence tomograph and standard automated perimetry,'' \emph{Journal
  of glaucoma}, vol.~21, no.~1, p.~49, 2012.

\bibitem{chapelle2009semi}
O.~Chapelle, B.~Scholkopf, and A.~Zien, ``Semi-supervised learning (chapelle,
  o. et al., eds.; 2006)[book reviews],'' \emph{IEEE Transactions on Neural
  Networks}, vol.~20, no.~3, pp. 542--542, 2009.

\bibitem{he2016deep}
K.~He, X.~Zhang, S.~Ren, and J.~Sun, ``Deep residual learning for image
  recognition,'' in \emph{Proceedings of the IEEE conference on computer vision
  and pattern recognition}, 2016, pp. 770--778.

\bibitem{wang2018weakly}
X.~Wang, H.~Chen, C.~Gan, H.~Lin, Q.~Dou, Q.~Huang, M.~Cai, and P.-A. Heng,
  ``Weakly supervised learning for whole slide lung cancer image
  classification,'' \emph{Medical Imaging with Deep Learning}, 2018.

\bibitem{zhou2016learning}
B.~Zhou, A.~Khosla, A.~Lapedriza, A.~Oliva, and A.~Torralba, ``Learning deep
  features for discriminative localization,'' in \emph{Proceedings of the IEEE
  Conference on Computer Vision and Pattern Recognition}, 2016, pp. 2921--2929.

\end{thebibliography}

\end{document}